%% file: main.tex
\documentclass{article}
\usepackage[nonatbib,final]{neurips_2022}
\usepackage[numbers]{natbib}
\usepackage[utf8]{inputenc} 
\usepackage[T1]{fontenc}    
\usepackage{hyperref}       
\usepackage{url}            
\usepackage{booktabs} 
\usepackage{amsfonts}       
\usepackage{nicefrac}       
\usepackage{microtype}      

\usepackage{mdframed}
\usepackage{amsmath,mathrsfs,amssymb,mathtools,bm}
\usepackage{scalerel}
\usepackage{amsthm}
\usepackage{txfonts}
\usepackage{fontawesome}
\usepackage{wrapfig}
\usepackage{todonotes}
\usepackage{xcolor}
\usepackage{setspace}
\usepackage{multirow}
\usepackage{bm,bbm}
\usepackage{paralist}
\usepackage{enumitem}
\usepackage{caption}[small]
\usepackage{subcaption} 

\usepackage[ruled,linesnumbered,noresetcount,vlined]{algorithm2e}

\newtheorem{problem*}{Problem}

\newtheorem*{example*}{Example}

\input{inputs.tex}

\usepackage{neurips_2022}

\usepackage[utf8]{inputenc} 
\usepackage[T1]{fontenc}    
\usepackage{hyperref}       
\usepackage{url}            
\usepackage{booktabs}       
\usepackage{amsfonts}       
\usepackage{nicefrac}       
\usepackage{microtype}      
\usepackage{xcolor}         

\title{Learning Constrained Optimization with Deep Augmented Lagrangian Methods}

\author{
   James Kotary\\
   Department of Computer Science\\
   University of Virginia\\
   \texttt{jk4pn@virginia.edu} 
  \And
   Ferdinando Fioretto\\
   Department of Computer Science\\
   University of Virginia\\
   \texttt{fioretto@virginia.edu} \\
}

\begin{document}

\maketitle

\begin{abstract}
Learning to Optimize (LtO) is a problem setting in which a machine learning (ML) model is trained to emulate a constrained optimization solver. Learning to produce optimal and feasible solutions subject to complex constraints is a  difficult task, but is often made possible by restricting the input space to a limited distribution of related problems.
Most LtO methods focus on directly learning solutions to the primal problem, and applying correction schemes or loss function penalties to encourage feasibility. This paper proposes an alternative approach, in which the ML model is trained instead to predict dual solution estimates directly, from which primal estimates are constructed to form dual-feasible solution pairs. This enables an end-to-end training scheme is which the dual objective is maximized as a loss function, and solution estimates iterate toward primal feasibility, emulating a Dual Ascent method. First it is shown that the poor convergence properties of classical Dual Ascent are reflected in poor convergence of the proposed training scheme. Then, by incorporating techniques from practical Augmented Lagrangian methods, we show how the training scheme can be improved to learn highly accurate constrained optimization solvers, for both convex and nonconvex problems.
\end{abstract}

\section{Introduction}

A substantial literature has been dedicated in recent years to the use of machine learning (ML) to accelerate the solution of optimization problems \cite{kotary2021end}. This research direction, often termed Learning to Optimize (LtO), aims to develop real-time constrained optimization capabilities, for applications requiring complex decisions to be made under stringent time constraints. These capabilities are increasingly demanded in settings such as job scheduling in manufacturing \cite{kotary2022fast}, power grid operation \cite{fioretto2020lagrangian}, and optimal control \cite{sambharya2023end}. Many approaches within the LtO scope aim at assisting external optimization solvers with information such as learned heuristics \cite{khalil2016learning} and active constraint prediction \cite{misra2022learning}. This paper falls into the category of \emph{end-to-end} LtO approaches, which train deep neural networks (DNNs) as \emph{proxy solvers} that estimate optimal solutions directly \cite{fioretto2020lagrangian,donti2021dc3}.

While learning to emulate a constrained optimization solver over arbitrary problem instances may not be feasible, highly accurate predictors can be obtained in many practical settings which require real-time optimization only over a distribution of closely-related problem instances. Still, feasibility of the learned proxy solver requires satisfying arbitarily complex constraints in the solutions output from a deep neural network (DNN), which is nontrivial to achieve. Most recent approaches train the DNN proxy solver to predict primal solutions, along with a correction scheme to minimize constraints violations \cite{donti2021dc3}, or with loss functions which incorporate constraint violation penalties \cite{fioretto2020lagrangian, park2023self}.

This paper proposes an altogether different approach, inspired by classical methods for dual optimization. While maintaining primal feasibility is in general difficult, constraints of the \emph{dual} problem are typically much easier to satisfy \cite{boyd2011distributed}. This motivates a novel LtO approach in which the DNN proxy solver is trained to estimate feasible dual solutions, from which associated primal solutions are recovered by the stationarity condition. This dual solution estimator is then trained to maximize the dual objective function, by which it iterates toward primal feasibility in a scheme which mimics the classical Dual Ascent. 
However, while dual ascent methods are theoretically appealing, they are often marred by poor convergence properties. This paper shows that these convergence challenges are inherited by end-to-end learning schemes for dual optimization. 
Thus, in this paper, we introduce a novel strategy 
that integrates transformations from practical Augmented Lagrangian Methods into the primal problem. This leads to end-to-end deep dual ascent methods with enhanced convergence properties. 
The paper showcases the ability of the proposed method to train neural networks as lightweight emulators of constrained optimization with remarkable accuracy, using both convex and nonconvex benchmark problems.

\section{Problem Overview}
\label{sec:problem_setting}

Consider a generic optimization problem with continuous variables, subject to equality and inequality constraints, all parameterized by a vector of coefficients $\bm{c} \in \mathbb{R}^c$. From this we may  define a mapping from any instance of coefficients $\bm{c}$ to the resulting optimal solution $\bm{x}^{\star}(\bm{c}) \in \mathbb{R}^n$:
\begin{subequations}
\label{eq:opt_parametric}
\begin{align}
    \label{eq:opt_parametric_objective}
    \bm{x}^{\star}(\bm{c}) \in \argmin_{\bm{x}} & \;\; f_{\bm{c}}(\bm{x}) \\
    \label{eq:opt_parametric_equalities}
    \textit{s.t.} \;\;\;& \;\; h_{\bm{c}}(\bm{x}) = \bm{0} \\
    \label{eq:opt_parametric_inequalities}
                 & \;\;g_{\bm{c}}(\bm{x}) \leq \bm{0} 
\end{align}
\end{subequations}
in which any choice of $\bm{c}$ specifies an optimization problem by determining functions the $f_{\bm{c}}: \mathbb{R}^n \to \mathbb{R}$, $g_{\bm{c}}: \mathbb{R}^n \to \mathbb{R}^m$, and $h_{\bm{c}}: \mathbb{R}^n \to \mathbb{R}^p$. In turn, this determines a corresponding optimal solution $\bm{x}^{\star}(\bm{c})$. 

The goal is to train an optimization \emph{proxy solver} $\cX_{\bm{\theta}}: \mathbb{R}^c \to \mathbb{R}^n$, over a distribution of problem parameters $\bm{c} \sim \cC$, which approximates the mapping $\bm{x}^{\star}(\bm{c})$ as defined by equation \eqref{eq:opt_parametric}. The proxy model $\cX_{\bm{\theta}}$ may consist of a  deep neural network $\cN_{\bm{\theta}}$ with \emph{trainable} weights $\bm{\theta}$, possibly combined with a \emph{non-trainable correction routine}  $\cS$ to improve solution quality, so that $\cX_{\bm{\theta}} = \cS \circ \cN_{\bm{\theta}}$. One may define the following as an ideal training goal for the optimization proxy model $\cX_{\bm{\theta}}$:
\begin{subequations}
\label{eq:ERM}
\begin{align}
    \min_{\bm{\theta}}  \;\; \underset{\bm{c} \sim \cC}{\mathbb{E}}\;\; \Big[ & \;\;f_{\bm{c}} \left( \, \cX_{\bm{\theta}}(\bm{c}) \, \right) \;\; \Big] \label{eq:ERM_objective} \\
    \textit{s.t.} \;\;\;& \;\; h_{\bm{c}}\left( \, \cX_{\bm{\theta}}(\bm{c}) \, \right) = \bm{0} \;\;\;\; \forall \bm{c} \sim \cC \label{eq:ERM_equalities}\\
                 & \;\;g_{\bm{c}}\left( \, \cX_{\bm{\theta}}(\bm{c}) \, \right) \leq \bm{0} \;\;\;\; \forall \bm{c} \sim \cC.
                 \label{eq:ERM_inequalities}
\end{align}
\end{subequations}

The training goal \eqref{eq:ERM} emphasizes that each solution $\cX_{\bm{\theta}}(\bm{c})$ produced by the proxy solver must be feasible to the constraints of problem \eqref{eq:opt_parametric}. Subject to these constraints on each output, their mean objective value should thus be minimized. Approaches to realizing this training goal may be separated into two main categories as follows. If feasibility could be \emph{ensured} in the outputs of an end-to-end trainable proxy model $\cX_{\bm{\theta}}$ for any DNN weights $\bm{\theta}$, the training goal \eqref{eq:ERM} could be realized by simply minimizing the loss function $f_{\bm{c}} \left( \, \cX_{\bm{\theta}}(\bm{c}) \, \right)$ of \eqref{eq:ERM_equalities}. This \emph{end-to-end} approach requires: {\textbf(1)} that the correction routine $\cS$  be differentiable for end-to-end training of $\cX_{\bm{\theta}} = \cS \circ \cN_{\bm{\theta}}$, and also {\textbf(2)} that feasibility of $(\cS \circ \cN_{\bm{\theta}})(\bm{c})$ is ensured for any DNN output $\cN_{\bm{\theta}}(\bm{c})$.   The main alternative is to treat $\cS$  as a post-processing step on the outputs of $\cN_{\bm{\theta}}$, as a separate stage at inference time rather than in end-to-end training. Some related work on each approach is reviewed next.

\section{Related Work}
\paragraph{End-to-End Learning with Feasibility Guarantees} Feasibility in the output space of an end-to-end trainable model $\cX_{\bm{\theta}} = \cS \circ \cN_{\bm{\theta}}$ is generally guaranteed by one of two approaches. A \emph{differentiable projection} $\cS$ onto the feasible set can be combined with a prediction model $\cN_{\bm{\theta}}$ to ensure feasibility of $\cX_{\bm{\theta}}$ in end-to-end training \cite{wilder2018melding}. However, this is generally inefficient except in limited cases where the constraints have special structure admitting a fast projection \cite{beck2017first}. To overcome this, a notable alternative to projections is proposed in \cite{donti2021dc3}, which instead corrects feasibility by unrolling differentiable gradient descent steps on the constraint violations of a solution estimate during training. Another strategy is based on \emph{reparametrization} of the model's output space to exclude infeasible solutions. However, this has only been demonstrated in cases where the constraints  (\eqref{eq:opt_parametric_equalities}, \eqref{eq:opt_parametric_inequalities}) are linear or quadratic inequalities \cite{frerix2020homogeneous,konstantinov2024new}. No known work has shown that feasibility to arbitrary constraints in the outputs of an end-to-end trainable model can be efficiently and reliably guaranteed. 

\paragraph{Learning with Soft Constraints} Several works  propose to satisfy constraints (\ref{eq:opt_parametric_equalities}, \ref{eq:opt_parametric_inequalities}) only approximately in the outputs of a DNN model $\cN_{\bm{\theta}}$, by training it with a loss function that incorporates penalties on the violation of each constraint \cite{Neuromancer2023}. At inference time, the outputs of  $\cN_{\bm{\theta}}$ are post-processed with a feasibility correction step $\cS$. This approach allows $\cS$ to take the form of a more lightweight model, such as a few iterations of Newton's method on the constraint violations \cite{kotary2023predict}, since the outputs of $\cN_{\bm{\theta}}$ are already nearly feasible. However, from this approach arises the problem of finding appropriate penalty weights. One line of work \cite{fioretto2020lagrangian,park2023self,kotary2022fast}  interprets the penalty weights as dual variables. Updating the penalty weights by update rules inspired by dual optimization improves feasibility, but their inexactness  leads to inconsistent constraint satisfaction. This often leads them to rely on hyperparameter search over initial values and update steps for the penalty weights.  

\paragraph{Learning Dual Solutions} Recently, another related direction has considered learning dual solutions to parametric optimization problems. The first such known work \cite{qiu2023dual} learns dual solutions to a convex conic programming model of optimal power flow. Two further works \cite{klamkin2024dual} and \cite{tanneau2024dual} have proposed to learn dual solutions to general linear programs and convex conic programs respectively, by incorporating a dual interior point method an a fast projection onto the dual cone, respectively. In these works, the learned model employs feasibility dual predictions as fast estimates on lower bounds of the primal objective function, as certificates of optimality.

\section{Preliminaries: Lagrangian Duality}
The method for learning constrained optimization  proposed in this paper is based on concepts of Lagrangian Duality. This section provides requisite background. The following notions are defined with respect to the parametric optimization \eqref{eq:opt_parametric}, for some particular instantiation of the parameters $\bm{c}$.

A \emph{Lagrangian function} returns weighted combinations of its objective and constraint functions:
\begin{equation}
\label{eq:lagrangian}
     \cL_{\bm{c}}(\bm{x},\bm{\lambda},\bm{\nu}) = f_{\bm{c}}(\bm{x}) + \bm{\lambda}^T g_{\bm{c}}(\bm{x}) +  \bm{\nu}^T h_{\bm{c}}(\bm{x}) , 
\end{equation}

and the Lagrangian Dual function is defined by a partial (unconstrained) minimization over the primal variables $\bm{x}$. The multipliers $\bm{\lambda},\bm{\nu}$ are called the \emph{dual variables}:
\begin{equation}
\label{eq:lagrangian_dual}
     d_{\bm{c}}(\bm{\lambda},\bm{\nu}) = \min_{\bm{x}} \cL_{\bm{c}}(\bm{x},\bm{\lambda},\bm{\nu}).
\end{equation}

The Lagrangian Dual Problem is to maximize the dual function, subject to that the inequality mutipliers $\bm{\lambda}$ are positive:
\begin{equation}
\label{eq:dual_problem}
    \argmax_{\bm{\lambda}, \bm{\nu}} \;\; d_{\bm{c}}(\bm{\lambda},\bm{\nu})  \;\;\textit{s.t.} \;\; \bm{\lambda} \geq 0.
\end{equation}

In cases when Strong Duality holds, solving the dual problem is equivalent to solving the original Primal problem \eqref{eq:opt_parametric}. If the pair $(\bm{\lambda}^{\star}, \bm{\nu}^{\star})$ solves the Dual problem \ref{eq:dual_problem}, then the solution $\bm{x}^{\star}$ to the Primal problem can be recovered via the stationarity condition
\begin{equation}
\label{eq:stationarity}
     \bm{x}^{\star} = \argmin_{\bm{x}} \cL_{\bm{c}}(\bm{x},\bm{\lambda}^{\star},\bm{\nu}^{\star}).
\end{equation}
This amounts to an \emph{unconstrained minimization}, which may or may not admit a closed-form solution. 

\section{Deep Dual Ascent}
\label{sec:deep_dual_ascent}

This section proposes a preliminary method for training proxy solvers of parametric constrained optimization problems \eqref{eq:opt_parametric}. Its main concept is to train a DNN model to estimate the solution of its \emph{dual} problem. For any observation of input parameters $\bm{c}$, a DNN model $\cN_{\bm{\theta}}$ with weights $\bm{\theta}$ is trained to produce estimated dual solutions $\left(\; \bm{\lambda}_{\bm{\theta}}(\bm{c}), \bm{\nu}_{\bm{\theta}}(\bm{c}) \;\right) = \cN_{\bm{\theta}}(\bm{c})$. The following training goal for our dual optimization proxy model is derived as a variant of the ideal training goal \eqref{eq:ERM}, but with respect to the dual problem \eqref{eq:dual_problem}:
\begin{subequations}
\label{eq:ERM_dual}
\begin{align}
\label{eq:ERM_dual_objective}
    \max_{\bm{\theta}}  \;\; \underset{\bm{c} \sim \cC}{\mathbb{E}}\;\; \Big[ & \;\;d_{\bm{c}} \left( \, \bm{\lambda}_{\bm{\theta}}(\bm{c}), \bm{\nu}_{\bm{\theta}}(\bm{c}) \, \right) \;\; \Big] \\
    \label{eq:ERM_dual_constraint}
    \textit{s.t.} \;\;\;& \;\; \bm{\lambda}_{\bm{\theta}}(\bm{c}) \geq \bm{0} \;\;\;\; \forall \bm{c} \sim \cC.
\end{align}
\end{subequations}

The key advantage of this choice is that the variables of the dual problem are subject only to nonnegativity constraints \eqref{eq:ERM_dual_constraint}, which are efficiently maintained in end-to-end training by applying a ReLU activation on the estimates $\bm{\lambda}_{\bm{\theta}}(\bm{c})$. This is equivalent to their Euclidean projection onto the positive orthant \cite{beck2017first}. Thus as described in Section \ref{sec:problem_setting}, the training goal \eqref{eq:ERM_dual} can be realized simply by training $\cN_{\bm{\theta}}$ to maximize the Lagrangian dual function $d_{\bm{c}}$ of equations \eqref{eq:lagrangian_dual} and \eqref{eq:ERM_dual_objective} as its loss function. Then, when solutions to the primal problem \eqref{eq:opt_parametric} are needed at test time, they are recovered from the predicted dual solutions $\left(\; \bm{\lambda}_{\bm{\theta}}(\bm{c}), \bm{\nu}_{\bm{\theta}}(\bm{c}) \;\right)$ by solving the stationarity condition \eqref{eq:stationarity}. In the general case, this amounts to an efficient \emph{unconstrained minimization}. In many important cases, this minimization can also be precomputed in closed form. Denoting as $D_{\bm{c}}(\bm{\lambda}, \bm{\nu})$ the function \eqref{eq:stationarity} which returns associated primal solutions from dual ones, the composite \emph{primal} proxy model is then
\begin{equation}
\label{eq:primal_prediction}
    \cX_{\bm{\theta}}(\bm{c}) = \left( \; D_{\bm{c}} \circ \cN_{\bm{\theta}} \;\right)(\bm{c})\;.
\end{equation}
\subsection{Training Algorithm} In order to maximize the Lagrangian dual loss function \eqref{eq:ERM_dual_objective} by gradient descent, its gradients with respect to the dual variables $\left(\; \bm{\lambda}_{\bm{\theta}}(\bm{c}), \bm{\nu}_{\bm{\theta}}(\bm{c}) \;\right)$ are required. The following textbook results form the basis of several classic algorithms for dual optimization \cite{boyd2011distributed}: 
\begin{subequations}
\label{eq:dual_function_gradients}
\begin{align}
    &\nabla_{\bm{\lambda}} d_{\bm{c}}(\bm{\lambda},\bm{\nu}) = g_{\bm{c}}(\bm{x}^{\star})\;, \label{eq:grad_lambda} \\
    &\nabla_{\bm{\nu}} d_{\bm{c}}(\bm{\lambda},\bm{\nu}) = h_{\bm{c}}(\bm{x}^{\star})\;,     \label{eq:grad_nu} \\
    \textit{where}  \;\;\;\;&    \bm{x}^{\star} = \argmin_x \; \cL_{\bm{c}}(x,\bm{\lambda},\bm{\nu}).
    \label{eq:where_lagrangian}
\end{align}
\end{subequations}
The Lagrangian function of  \eqref{eq:where_lagrangian} is as defined as in \eqref{eq:lagrangian}. These derivative rules enable a training scheme aimed at realizing \eqref{eq:ERM_dual} directly. For reasons described below, it is named \emph{Deep Dual Ascent} (DDA).

\begin{algorithm}[H]
  {\small
    \DontPrintSemicolon
    \caption{\!Deep Dual Ascent \!\!\!\!\!\!\!\!\!\!\!\!\!\!}
    \label{alg:DDA}
    \setcounter{AlgoLine}{0}
    \SetKwInOut{Input}{input}
  
    \Input{$\{\bm{c}_{(i)}\}_{i=1}^N$: Input data, $\cN_{\bm{\theta}}$: a deep neural network, $\alpha$: the learning rate}
      \For{$i = 1$ to $N$}
     {
       $\hat{\bm{\lambda}}, \hat{\bm{\nu}} \gets \cN_{\bm{\theta}}(\bm{c}_{(i)})$ \label{line:predict}\\
     $\hat{\bm{\lambda}} \gets \textit{ReLU}(\hat{\bm{\lambda}} )$ \label{line:relu}\\
      $\bm{x}^{\star} \gets \argmin_x f_{\bm{c}_{(i)}}(\bm{x}) +  \hat{\bm{\lambda}}^T \bm{g}_{\bm{c}_{(i)}}(\bm{x}) + \hat{\bm{\nu}}^T h_{\bm{c}_{(i)}}(\bm{x})$ \label{line:getprimal}\\
      $\nabla_{\bm{\lambda},\bm{\nu}} \;d \gets \left[ h_{\bm{c}_{(i)}}(\bm{x}^{\star}),\; g_{\bm{c}_{(i)}}(\bm{x}^{\star})   \right]$ \label{line:getgradient}\\
      $\bm{g} \gets \nabla_{\bm{\theta}} d$ \text{via backpropagation of} $\nabla_{\bm{\lambda},\bm{\nu}} \;d$ \text{through} $\cN_{\bm{\theta}}$ \text{by automatic differentiation}\label{line:backprop}\\
      $\bm{\theta} \gets \bm{\theta} + \alpha \cdot \bm{g}$
      
    }
    \Return $\cN_{\bm{\theta}}$
  }
  \end{algorithm}
Algorithm \ref{alg:DDA} outlines a single iteration of Deep Dual Ascent in stochastic gradient ascent mode. It trains a DNN model to directly predict estimated solutions to the dual problem associated to the underlying primal problem \eqref{eq:opt_parametric}. Lines \ref{line:predict} and \ref{line:relu} represent the forward pass in training, emphasizing that feasibility of the inequality duals is maintained by a ReLU function. For the backward pass, the associated primal solution is first obtained by unconstrained optimization of the Lagrangian function as shown in line \ref{line:getprimal}. Then, it is used to compute gradients of the dual objective function by applying the formulas \eqref{eq:dual_function_gradients} as shown in line \ref{line:getgradient}. Finally, those gradients are further backpropagated to the weights of the dual prediction model to complete a gradient ascent update.

\subsection{Dual Ascent Interpretation} The use of dual function gradients \eqref{eq:dual_function_gradients} is fundamental to many algorithms which solve optimization problems via their dual problems. In fact, the behavior of Algorithm \ref{alg:DDA} is best understood in light of its similarities to the classical Dual Ascent method.  

Classical Dual Ascent  solves the dual problem \eqref{eq:dual_problem}, in order to reach a solution to the primal problem \eqref{eq:opt_parametric} by applying the transformation \eqref{eq:stationarity}. In the absence of inequality constraints \eqref{eq:opt_parametric_inequalities}, it amounts to gradient ascent on the unconstrained Lagrangian dual function. In the present case where both equalities and inequalities must be satisfied, the constrained dual problem \eqref{eq:dual_problem} is solved by projected gradient ascent, in which the projection onto $\bm{\lambda} \geq 0$ is trivially computed as the clamping function $\left[ \cdot \right]_+$, also known as ReLU. This variant is sometimes called Projected Dual Ascent \cite{boyd2011distributed}:
\begin{subequations}
\label{eq:dual_ascent}
\begin{align}
    x^{k} & = \argmin_x \cL_{\bm{c}}(x,\bm{\lambda}^k,\bm{\nu}^k) \label{eq:dual_ascent_getx}   \\
    \bm{\nu}^{k+1} & = \; \bm{\nu}^{k} + \alpha \cdot h_{\bm{c}}(x^k) \label{eq:dual_ascent_nu} \\
    \bm{\lambda}^{k+1} & = \left[ \bm{\lambda}^{k} + \alpha \cdot g_{\bm{c}}(x^k) \right]_+ \label{eq:dual_ascent_lambda}
\end{align}
\end{subequations}

where $h_{\bm{c}}(\bm{x}^k)$ and $g_{\bm{c}}(\bm{x}^k)$ are again recognized as the gradients of $d_{\bm{c}}$ with respect to the the dual variables $\bm{\nu}$ and $\bm{\lambda}$, respectively, and $\alpha$ is a stepsize parameter.

The main difference between Dual Ascent for solving the dual problem \eqref{eq:dual_problem}, and Deep Dual Ascent for training an optimization proxy model as prescribed by \eqref{eq:ERM_dual}, is as follows. Once gradients of the Lagrangian dual function $d_{\bm{c}}$ are calculated with respect to the (predicted) dual variables (Line \ref{line:getgradient}),                  they are back-propagated down to the weights $\bm{\theta}$ of the DNN model $\cN_{\bm{\theta}}$ which predicts them (Line \ref{line:backprop}). 
Gradient ascent steps are thereby applied to the underlying neural network weights, rather than the dual variables themselves as in equations \eqref{eq:dual_ascent_nu} and \eqref{eq:dual_ascent_lambda}. 

Like Dual Ascent, DDA maintains feasibility in its \emph{dual} solution estimates, while iterating toward \emph{primal} feasibility at convergence to the optimal dual function value. However, as shown in Section \ref{sec:Experiments}, Deep Dual Ascent suffers from extremely slow convergence in training, and is not practically viable for efficient training of optimization proxy models. Fortunately, this behavior is not surprising as it reflects the well-known convergence issues shared by classical Dual Ascent \cite{boyd2011distributed}. We find a natural path to improvement by incorporating concepts from the so-called Augmented Lagrangian Methods (ALM). The next section details how Deep Dual Ascent can be modified to yield training schemes for optimization proxies which are extremely efficient and reliable in practice.

\section{Deep Augmented Lagrangian Method}
To improve on its poor convergence properties \cite{boyd2011distributed}, the Dual Ascent method is often modified by augmenting the Lagrangian function with a penalty on the constraint residuals. In the special case of equality constrained problems (i.e. in the absence of \eqref{eq:opt_parametric_inequalities}), a Dual Ascent method which uses the following modified Lagrangian function where $\rho$ is a chosen penalty weight on the equality residuals:
\begin{equation}
\label{eq:lagrangian_aug}
     \cL_{\bm{c}}(x,\bm{\nu}) \coloneqq f_{\bm{c}}(\bm{x}) +   \bm{\nu}^T h_{\bm{c}}(\bm{x}) + \rho\|h_{\bm{c}}(\bm{x})\|^2
\end{equation}
yields the well-known Augmented Lagrangian Method (ALM), also called the Method of Multipliers. Its superior convergence properties lead it to be preferred over Dual Ascent for solving equality-constrained problems. Beyond the equality-constrained case, much work has been dedicated to extending the ALM to optimization problems \eqref{eq:opt_parametric} which also contain inequality constraints. For example, \cite{rockafellar1974augmented} applies an additional penalty to a smoothed function measuring the inequality constraint residuals. To improve on our Deep Dual Ascent training scheme, we take a different approach, inspired by techniques common to many practical large-scale optimization solvers.  

\subsection{Box-Constrained Reformulation}
To improve the convergence of Deep Dual Ascent, we form an adaptation an ALM for inequality-constrained optimization \eqref{eq:opt_parametric} which is perhaps best known as that employed by the LANCELOT package for large-scale nonlinear optimization \cite{conn1991globally}. Its main routine is based on a \emph{box-constrained} augmented Lagrangian function, to which the typical equality-constrained ALM method can be applied. The following generic form encompasses any inequality-constrained parametric optimization:
\begin{subequations}
\label{eq:lancelot_form}
\begin{align}
   \argmin_{x} \;\;\;& f_{\bm{c}}(\bm{x})  \\
   \textit{s.t.} \;\;\;&  \bm{l} \leq r_{\bm{c}}(\bm{x}) \leq \bm{u}
\end{align}
\end{subequations}

wherein equality constraints correspond to elements where $\bm{l}=\bm{u}$. This form can always be reformulated as follows, by the introduction of slack variables:
\begin{subequations}
\label{eq:lancelot_standard_form}
\begin{align}
   \argmin_{\bm{l} \leq \bm{x} \leq \bm{u} } \;\;\;& f_{\bm{c}}(\bm{x})  \\
   \textit{s.t.} \;\;\;&  h_{\bm{c}}(\bm{x}) = \bm{0} .
\end{align}
\end{subequations}

This problem can now be viewed as that of minimizing the function $f(\bm{x}) + \delta_{[l,u]}(\bm{x})$ subject to a nonlinear equality constraint, where $\delta_{[l,u]}$ is an indicator function taking value $0$ inside its box domain and $\infty$ elsewhere. Its standard form avoids explicit inequality constraints, along with associated dual variables $\bm{\lambda}$. The augmented Lagrangian function is now
\begin{equation}
\label{eq:lagrangian_box}
     \cL_{\bm{c}}(\bm{x},\bm{\nu}) = f_{\bm{c}}(\bm{x})+ \delta_{[l,u]}(\bm{x}) +  \bm{\nu}^T h_{\bm{c}}(\bm{x}) + \rho\|h_{\bm{c}}(\bm{x})\|^2.
\end{equation}
Evaluating the Lagrangian dual function and its gradients amounts to minimizing \eqref{eq:lagrangian_box} with respect to primal variables $x$. This now requires a \emph{box-constrained} minimization, rather than an unconstrained one. The resulting ALM routine can be viewed as Dual Ascent on the transformed problem:

\begin{subequations}
\label{eq:lancelot}
\begin{align}
    \label{eq:lancelot_objective}
    x^{k} & = \argmin_{l \leq x \leq u} f_{\bm{c}}(\bm{x})+  (\bm{\nu}^k)^T h_{\bm{c}}(\bm{x}) + \rho^k \|h_{\bm{c}}(\bm{x})\|^2   \\
    \label{eq:lancelot_constraint}
    \bm{\nu}^{k+1} & = \; \bm{\nu}^{k} + \rho^k \cdot h_{\bm{c}}(x^k).
\end{align}
\end{subequations}

The iterations \eqref{eq:lancelot} can be recognized as the core optimization routine used by LANCELOT for large-scale nonlinear optimization. Convergence of the LANCELOT algorithm to KKT points of \emph{nonconvex} programs was proven in \cite{conn1991globally} under regularity assumptions, and another variation of the method was shown to converge in \cite{andreani2008augmented} under a reduced set of conditions. Accordingly, Section \ref{sec:Experiments} demonstrates the use of the following proposed method to learn both convex and nonconvex optimization.

\paragraph{Box-Constrained Optimization Step} Of course, the routine \eqref{eq:lancelot} relies on the box-constrained optimization     \eqref{eq:lancelot_objective} being much more efficiently solvable than the overall problem \eqref{eq:lancelot_form}. In practice, its solution is generally warmstarted from that of the previous iteration, and need not be solved to a global optimum at every iteration for good performance. Popular algorithms for solving   \eqref{eq:lancelot_objective} include L-BGFS-B, a limited-memory Quasi-Newton method specialized for box constraints \cite{liu1989limited}. 

\subsection{Training Optimization Proxy Models with Deep ALM}

Based on these optimization strategies, we propose a \emph{Deep Augmented Lagrangian Method} (Deep ALM)  for training neural networks as proxy optimization solvers. It can be viewed as a variant of Deep Dual Ascent, applied to the reformulated problem \eqref{eq:lancelot_standard_form}, and using the augmented Lagrangian function \eqref{eq:lagrangian_box}. Due to its equality-constrained reformulation, the dual prediction model $ \bm{\nu}_{\bm{\theta}}(\bm{c})  = \cN_{\bm{\theta}}(\bm{c})$ produces only one set of (unconstrained) dual variable estimates. 

A single epoch of Deep ALM training by stochastic gradient ascent is prescribed in Algorithm     \ref{alg:DALM}. The box-constrained optimization of line \ref{line:box_constrained_opt_step} is implemented using the L-BGFS-B method of \texttt{scipy} \cite{virtanen2020scipy}. For each data sample, the solution $\bm{x}^{\star}$ of the previous epoch is used to hotstart its optimization in the next epoch, which greatly accelerates training. As in conventional ALM methods, the update rule for $\rho$ may be the subject of various design choices. In this article, we use a simple update rule in which $\rho$ is increased at each epoch by a constant factor $\gamma$.

\begin{algorithm}[H]
  {\small
    \DontPrintSemicolon
    \caption{\!Deep Augmented Lagrangian Method \!\!\!\!\!\!\!\!\!\!\!\!\!\!}
    \label{alg:DALM}
    \setcounter{AlgoLine}{0}
    \SetKwInOut{Input}{input}
  
    \Input{$\{\bm{c}_{(i)}\}_{i=1}^N$: Input data, $\cN_{\bm{\theta}}$: a deep neural network, $\alpha$: the learning rate}
      \For{$i = 1$ to $N$}
     {
       $\hat{\bm{\nu}} \gets \cN_{\bm{\theta}}(\bm{c}_{(i)})$\\
      $\bm{x}^{\star} \gets \underset{\bm{l} \leq \bm{x} \leq \bm{u}}{\argmin} \;\;f_{\bm{c}_{(i)}}(\bm{x}) + \hat{\bm{\nu}}^T h_{\bm{c}_{(i)}}(\bm{x}) + \rho \| h_{\bm{c}_{(i)}}(\bm{x}) \|^2$ \label{line:box_constrained_opt_step}\\

      $\nabla_{\bm{\nu}} \;d \gets  h_{\bm{c}_{(i)}}(\bm{x}^{\star})  $ \label{line:getgradient_DALM}\\
      $\bm{g} \gets \nabla_{\bm{\theta}} d$ \text{via backpropagation of} $\nabla_{\bm{\nu}} \;d$ \text{through} $\cN_{\bm{\theta}}$ \text{by automatic differentiation}\\
      $\bm{\theta} \gets \bm{\theta} + \alpha \cdot \bm{g}$

       $\rho \gets \text{Update}(\rho)$
    }
    \Return $\cN_{\bm{\theta}}$
  }
  \end{algorithm}

\section{Experiments}
\label{sec:Experiments}

We evaluate the ability of both Deep Dual Ascent and Deep ALM to learn to solve parametric \emph{convex} and \emph{nonconvex} optimization problems. Following prior works \cite{donti2021dc3} and \cite{park2023self}, we first evaluate their ability to learn Quadratic Programming (QP) problems, followed by a variation in which the linear objective term is replaced by a nonconvex sinusoidal objective.

\paragraph{Convex QP.} A parametric QP is defined in standard form as a function of its cost coefficients:
\begin{subequations}
\label{eq:QP}
\begin{align}
   \bm{x}^{\star}(\bm{c}) =   \argmin_{\bm{x}} \;\;&\;\;\; \bm{x}^T \bm{Q} \bm{x} + \bm{c}^T \bm{x} \\
    \textit{s.t.} & \;\;\;\;\;\;\; \bm{A}\bm{x} = \bm{b} \\
                & \;\;\;\;\;\;\;\;\;\; \bm{x} \geq \bm{0},
\end{align}
\end{subequations}
where $\bm{A} \in \mathbb{R}^{p \times n}$, $\bm{b} \in \mathbb{R}^{p}$ and $\bm{Q} \in \mathbb{R}^{n \times n}$ are uniform randomly generated with $p=20$ and $n=50$. That is, a $50$ dimensional variable is optimized subject to $50$ inequality and $20$ equality constraints.  

\paragraph{Nonconvex Variant} An important question is whether Deep ALM has the ability to learn \emph{nonconvex} optimization. To test this, the follow variant of problem \eqref{eq:QP} is posed:
\begin{subequations}
\label{eq:nonconvex_variant}
\begin{align}
   \bm{x}^{\star}(\bm{c}) =   \argmin_{\bm{x}} \;\;&\;\;\; \bm{x}^T \bm{Q} \bm{x} + \bm{c}^T \sin (\bm{x}) \\
    \textit{s.t.} \;\;& \;\;\;\;\;\;\; \bm{A}\bm{x} = \bm{b} \\
                & \;\;\;\;\;\;\;\;\;\; \bm{x} \geq \bm{0}.
\end{align}
\end{subequations}
The object of each learning task is to learn $\bm{x}^{\star}$ as a function of $\bm{c}$ in problems \eqref{eq:QP} and \eqref{eq:nonconvex_variant}. A dataset of $10,000$ instances $\bm{c}_i \sim \cC$ randomly generated with each component uniformly sampled from $\left[ -20, 20 \right]$. Training and test data are split $80:20$.

\subsection{Performance Metrics}
The performance of each method is evaluated in terms of the following criteria each at epoch of training, where all reported results are taken over the test set:
\begin{itemize}[leftmargin=*, parsep=2pt, itemsep=0pt, topsep=0pt]
    \item Dual function optimality gap: measures suboptimality of the Lagrangian dual loss function. Computed as $d_{\bm{c}}(\bm{\lambda}^{\star},\bm{\nu}^{\star}) - d_{\bm{c}}(\bm{\lambda},\bm{\nu})$ where $(\bm{\lambda}^{\star},\bm{\nu}^{\star})$ are the true optimal dual values.
    \item Primal objective value $f_{\bm{c}}(\bm{x})$. For reference, the mean optimal objective value $f_{\bm{c}}(\bm{x}^{\star})$ is also shown. Since the primal optimality gap changes from negative to positive, the nominal objective value is more interpretable and thus included instead.
    \item Residual of the equality constraint, computed as $\| h_{\bm{c}}(\bm{x}) \|_2$. Since primal solutions must be feasible, this is the most important convergence metric for Deep ALM.
    \item Residual of the inequality constraint $\|\; \left[ g_{\bm{c}}(\bm{x}) \right]_+ \|_2$. Not included in the case of Deep ALM, since residuals are kept at zero via box-constrained optimization.
    \item Solution residual: $L2$ error w.r.t. the true, precomputed optimal solutions obtained from \texttt{cvxpy} \cite{diamond2016cvxpy} in the case of the convex QP \eqref{eq:QP} and \texttt{cyipopt} \cite{biegler2009large} in the case of the nonconvex program \eqref{eq:nonconvex_variant}. Computed as $\| \bm{x} - \bm{x}^{\star} \|_2$.

\end{itemize}

\subsection{Parameters and Settings}

The dual solution prediction network $\cN_{\bm{\theta}}$ in the following experiments is a five-layer feedforward ReLU network. Deep ALM performs best when trained by SGD optimizer with a low learning rate; results are reported using rate $1e-5$. Deep Dual Ascent performs best when paired with Adam optimizer, and results are reported using learning rate $5e-4$. Xavier initialization is applied, along with batch normalization. Each method is run for 200 epochs of training. For Deep ALM, $\rho$ is initialized at $10.0$ and increased per epoch at an exponential rate of $\gamma = 1.05$. Batch size is $50$.

\subsection{Results}

Solid curves represent the mean value of each metric, and shaded regions represent the standard deviation over the test set. Note that when presented in log scale, the regions of standard deviation skew downward as an effect of the y-axis scaling.

\begin{figure*}
    \centering
    \vspace{-2pt}
     \includegraphics[width=0.49\textwidth,trim={0.4in 1.3in 0.8in 1.3in},clip]{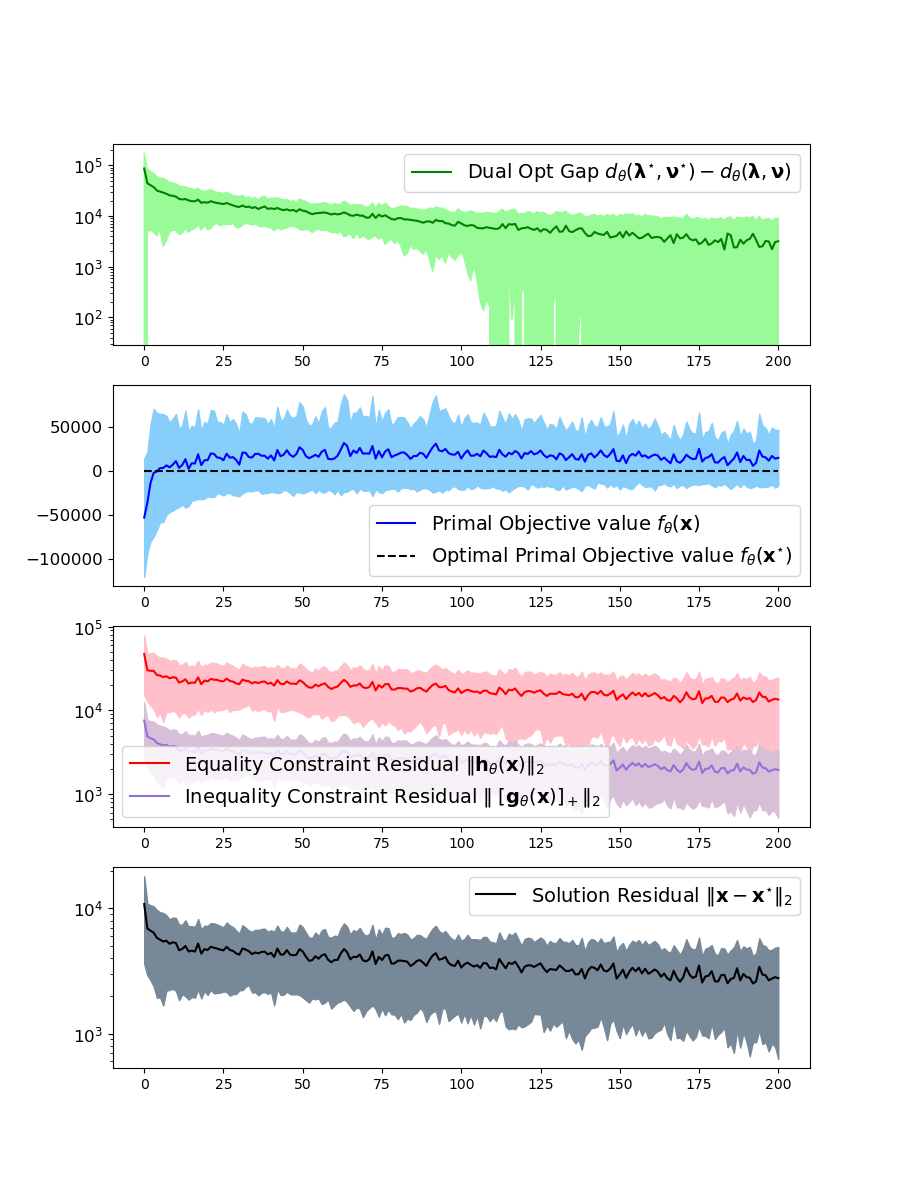}
     \includegraphics[width=0.49\textwidth,trim={0.4in 1.3in 0.8in 1.3in},clip]{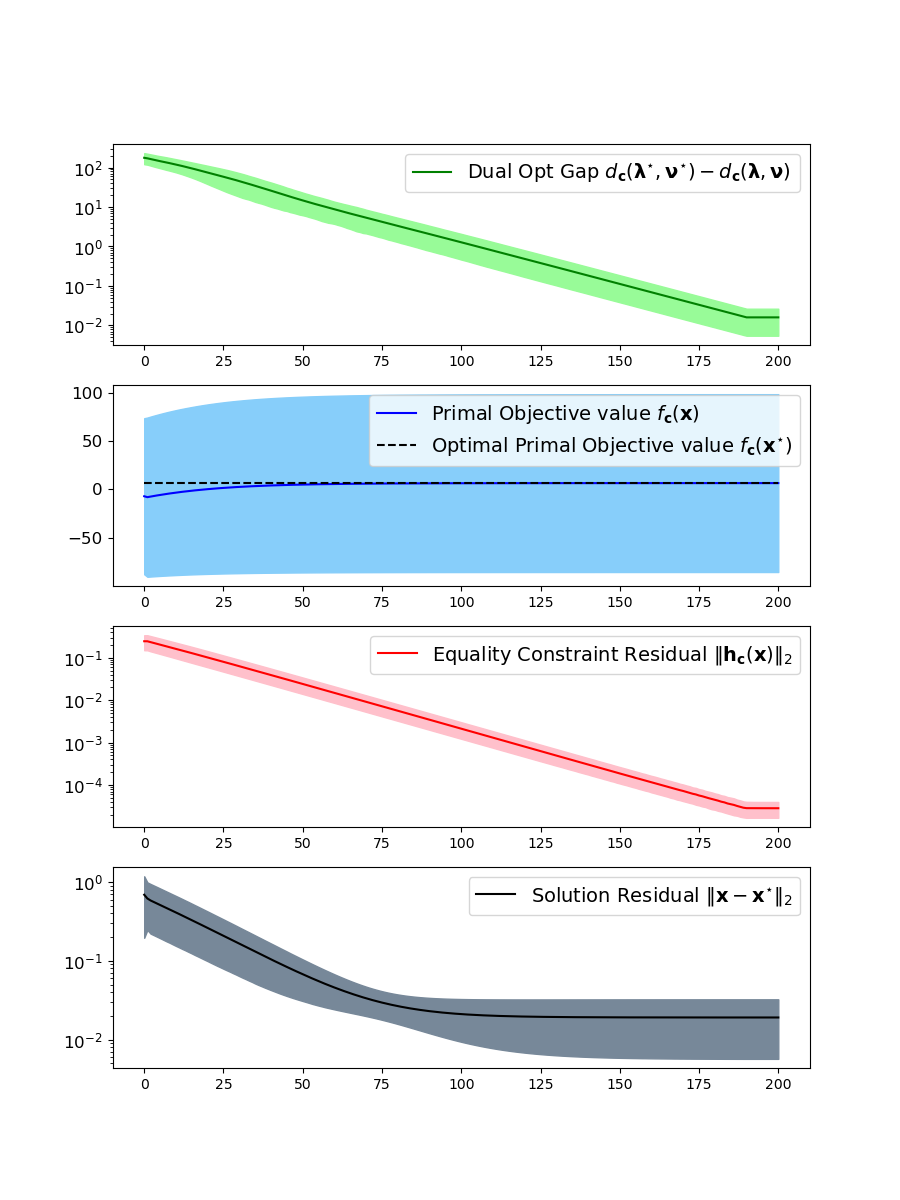}
    \caption{Training with Deep Dual Ascent (left) and Deep ALM (right) to solve the convex QP problem \eqref{eq:QP}. Mean and standard deviation over the test set shown by bold curves and shaded regions.}
    \label{fig:results_convex_QP_DALM}
\end{figure*}

Figure \ref{fig:results_convex_QP_DALM} shows the mean and standard deviation of each performance metric over the test set of problem instances. The baseline DDA method is shown at left, for comparison to Deep ALM on the right. The dual optimality gap, shown in green, corresponds to suboptimality of the dual loss function used in training. Minding the difference between y-axis scales between left and right, note the improvement of several orders of magnitude between DDA and Deep ALM. 

This difference in optimality of the learned dual solutions is reflected in the accuracy of the resulting primal solutions. Since primal-feasible solutions are desired, the equality constraint residual shown in red is most salient as a measure of convergence. Remarkably, Deep ALM training erases the equality constraint residual to nearly \emph{five decimal places of precision} on average, with low variation over the test set. This coincides with nearly zero gap in the primal objective, shown in blue. 

Note that despite being a minimization problem, the primal objective value increases in training before crossing from super-optimal to slightly sub-optimal on average. This behavior is typical in dual optimization. Note additionally the variation in optimal objective values over the test set, indicating that Deep ALM has learned to solve a nontrivial distribution of optimization instances with highly variable solutions.

Finally, Euclidean distance from the precomputed primal solution is shown in black. Despite not being trained directly, it also reaches low error. Residual values from the precomputed optima are reduced to nearly $1e-2$, and in any case are not directly relevant to solution quality. 

\begin{figure*}
    \centering
    \vspace{-2pt}
     \includegraphics[width=0.65\textwidth,,trim={0.4in 1.3in 0.8in 1.3in},clip]{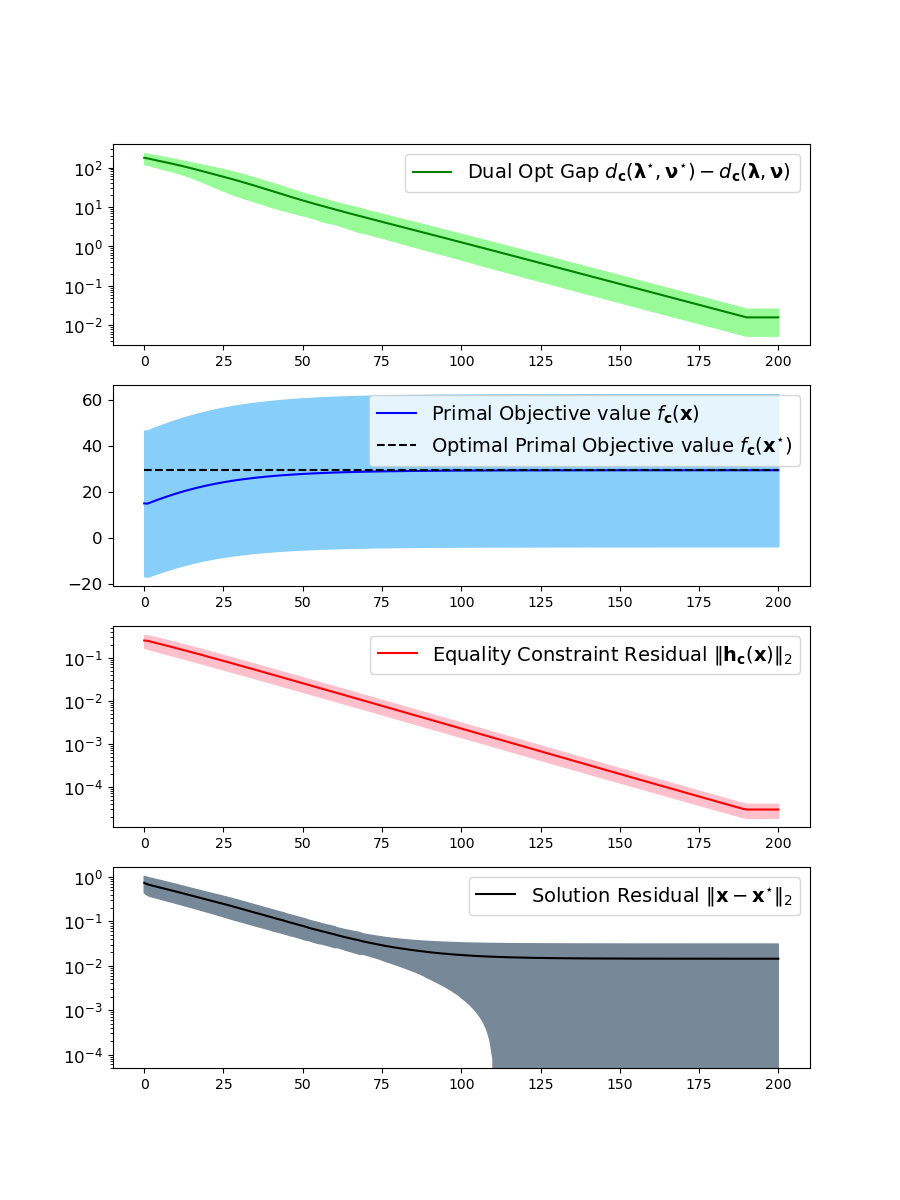}
    \caption{Training with Deep ALM to solve the nonconvex program \eqref{eq:nonconvex_variant}.}
    \label{fig:results_nonconvex_QP_DALM}
\end{figure*}

\paragraph{Learning Nonconvex Optimization.} Figure \ref{fig:results_nonconvex_QP_DALM} reports the same set of metrics while learning to solve the nonconvex optimization problem \eqref{eq:nonconvex_variant} with Deep ALM. Remarkably, similar accuracy is achieved as in the convex case, including equality constraint satisfaction up to five decimal places on average. This highlights the ability of Deep ALM to learn difficult optimization problems, using lightweight DNN models, with negligible error. 

\section{Conclusions}
This paper has proposed a novel system of learning to solve constrained optimization problems, based on classical methods for dual optimization. By reframing the learning to optimize problem in terms of its Lagrangian dual, we showed how end-to-end learning of proxy solvers can be enabled by efficiently maintaining dual feasibility during training. Poor convergence properties of this direct approach to dual optimization learning were explained by analogy to classical Dual Ascent, motivating an improved learning scheme based on Augmented Lagrangian methods. The resulting Deep ALM was demonstrated to achieve remarkable accuracy in learning to solve convex and nonconvex optimization problems, producing fast solutions with negligible loss to optimality and feasibility.

\section*{Acknowledgments}
This research is partially supported by NSF grants 2334936, 2334448, and NSF CAREER Award 2401285.
The views and conclusions of this work are those of the authors only.

\bibliographystyle{abbrvnat}
\bibliography{lib}
\newpage


\end{document}

%% file: inputs.tex
\usepackage{todonotes}
\usepackage{xcolor}
\usepackage{setspace}
\usepackage{multirow}
\usepackage{bm,bbm}
\usepackage{paralist}

\DeclareMathOperator*{\argmax}{argmax}
\DeclareMathOperator*{\argmin}{argmin}

\usepackage{amsmath}
\usepackage{multirow}
\usepackage{multicol}
\usepackage{url}

\definecolor{darkgreen}{RGB}{204,102,0}

\newcommand{\rev}[1]{{\color{purple}{#1}}
	\todo[caption={},color=purple!20!]
		{\begin{spacing}{0.625}
			{\footnotesize to review}
		\end{spacing}}
	}

\newcommand{\cC}{\mathcal{C}}

 \newcommand{\cL}{\mathcal{L}}
\newcommand{\cM}{\mathcal{M}} \newcommand{\cN}{\mathcal{N}}
 
 \newcommand{\cS}{\mathcal{S}}

 \newcommand{\cX}{\mathcal{X}}

\usepackage{esvect}

\makeatletter
\newcommand{\oset}[3][0ex]{\mathrel{\mathop{#3}\limits^{\vbox to#1{\kern-2\ex@\hbox{$\scriptstyle#2$}\vss}}}}
\makeatother

\usepackage{letltxmacro}
\LetLtxMacro\orgvdots\vdots
\LetLtxMacro\orgddots\ddots